# ATLO-ML: Adaptive Time-Length Optimizer for Machine Learning — Insights from Air Quality Forecasting

I-Hsi Kao and Kanji Uchino

*Abstract*— Accurate time-series predictions in machine learning are heavily influenced by the selection of appropriate input time length and sampling rate. This paper introduces ATLO-ML, an adaptive time-length optimization system that automatically determines the optimal input time length and sampling rate based on user-defined output time length. The system provides a flexible approach to time-series data pre-processing, dynamically adjusting these parameters to enhance predictive performance. ATLO-ML is validated using air quality datasets, including both GAMS-dataset and proprietary data collected from a data center, both in time series format. Results demonstrate that utilizing the optimized time length and sampling rate significantly improves the accuracy of machine learning models compared to fixed time lengths. ATLO-ML shows potential for generalization across various time-sensitive applications, offering a robust solution for optimizing temporal input parameters in machine learning workflows.

## I. INTRODUCTION

AIR quality is closely intertwined with the daily lives of individuals, not only impacting critical issues such as public health and livelihood but also presenting a significant barrier to economic development and societal progress. Research has confirmed that the primary causes of air pollution are urban population growth, industrial activities, and vehicular emissions [1, 2]. The quality of air substantially affects the overall health of a region's population [3, 4, 5]. There is a well-documented positive correlation between air pollution and mortality rates, with numerous epidemiological studies having established a link between air pollutants (such as PM or NO) and excess daily mortality [6] and morbidity [7]. Environmental regulatory authorities, as early as 1988, set maximum allowable concentrations for various air pollutants, including $SO_2$ (80 μg/m³), $NO_2$ (80 μg/m³), and RSPM (100 μg/m³) [8].

Air quality is a critical factor that impacts not only human health but also the performance and longevity of equipment. The controlled environment within these facilities is essential for maintaining the integrity and longevity of sensitive electronic equipment. Poor air quality can potentially lead to various issues, including corrosion of electronic components, reduced energy efficiency, and increased risk of hardware failures [9, 10].

These problems can result in significant operational disruptions and financial losses due to equipment damage, reduced performance, and potential downtime [11]. Studies have shown that unplanned downtime in manufacturing can produce extra cost [12], and companies that effectively manage operational risks can avoid in annual revenue losses from downtime [13]. Furthermore, equipment failures due to inadequate maintenance or environmental factors can lead to substantial production losses [14].

In data center operations, indoor air quality plays a vital role in ensuring efficient and reliable performance. The American Society of Heating, Refrigerating and Air-Conditioning Engineers (ASHRAE) has been at the forefront of this effort. Their Technical Committee 9.9 (TC 9.9) has published comprehensive guidelines [15, 16]. The International Organization for Standardization (ISO) has also contributed to this field with standards such as ISO 14644 [17], which defines cleanliness classes for air particles in cleanrooms and associated controlled environments. In addition to the ASHRAE and ISO guidelines, the ANSI/ISA-71.04-2013 standard [18], plays a crucial role in defining acceptable levels of corrosive contaminants for electronic equipment. Additionally, the European Commission's Code of Conduct on Data Centre Energy Efficiency includes recommendations on environmental control and air quality management as part of its best practices for energy-efficient data center operations [19].

Given the considerations, the management and prediction of indoor air quality are of paramount importance. In a study by Shen Yang [20], researchers developed a simulation toolbox using MATLAB App Designer, integrating it into healthy building design for indoor air quality forecasting. However, as with any simulation tool, the accuracy and reliability of predictions are significantly influenced by various parameters, including initial conditions, resolution settings, and other configurable variables. Consequently, manual parameter adjustment and optimization remain inevitable in the simulation and prediction process, necessitating a balance between automated processes and expert intervention to ensure optimal results.

The use of machine learning for measurement analysis has been a well-established technique for years [21, 22, 23, 24, 25, 26, 27]. To mitigate the challenges associated with excessive parameter adjustments and initial parameter configurations, machine learning and data-driven approaches for air quality

(Corresponding author: I. -H. Kao)

I. -H. Kao is with the Technology Strategy Unit, Fujitsu Research of America Inc., Santa Clara, CA 95054, USA (e-mail: ikao@fujitsu.com).

K. Uchino is with the Artificial Intelligence Laboratory, Fujitsu Research of America Inc., Santa Clara, CA 95054, USA (e-mail: kanji@fujitsu.com).

prediction have been extensively employed in numerous studies. For instance, recent research [28], such as that conducted at the eLUX lab in the University of Brescia, has demonstrated the efficacy of integrating sensor networks with artificial neural networks to predict and control indoor air quality parameters, particularly $CO_2$ concentration. Further illustrating this trend, a study [29] employed four machine learning models to predict indoor carbon dioxide levels: Ridge regression, Decision Tree (DT), Random Forest (RF), and Multi-Layer Perceptron. In a similar vein, another group of researchers [30] utilized six machine learning models for the same purpose: Support Vector Machine (SVM), AdaBoost, RF, Gradient Boosting, Logistic Regression, and Multi-Layer Perceptron. These studies collectively demonstrate the diverse approaches available in applying machine learning techniques to indoor $CO_2$ prediction, underscoring the versatility and potential of these methods in addressing complex air quality challenges.

$CO_2$, while significant, is not the sole indicator of indoor air quality. A more comprehensive approach to air quality assessment has been adopted in recent studies. For instance, [31] implemented a range of techniques including artificial neural networks, partial least squares, RF, and multiple linear regression to predict multiple air quality indices such as CO, $CO_2$, $NO_2$, radon, volatile organic compounds (VOCs), and semi-volatile organic compounds (SVOCs). In a different study, [32] utilized multiple linear regression, neural networks, and recurrent neural networks to predict PM10 and PM2.5 levels, considering variables like NO, $NO_2$, NOx, CO, $CO_2$, temperature, and humidity. Further expanding on this multi-faceted approach, [33] conducted a comprehensive comparative analysis of eight machine learning models in their ability to predict six distinct air quality indices.

While time series prediction methodologies have reached a considerable level of maturity [34, 35, 36], the optimal selection of input time length and sampling rate for various output time horizons remains an underexplored area. This paper focuses on analyzing and discussing the impact of input time length and sampling rate on prediction accuracy. The Adaptive Time-Length Optimizer for Machine Learning (ATLO-ML) represents a novel approach designed to dynamically adjust input time length and sampling rate to enhance the accuracy of time series predictions. The effectiveness of ATLO-ML was validated through experiments using both publicly available datasets, specifically GAMS-dataset [37], and proprietary data collected from the Data Center. The machine learning validation encompassed a diverse range of algorithms, including DT [38], K-Nearest Neighbors (KNN) [39, 40], LASSO regression [41], LightGBM [42], RF [43], SVM [44], and XGBoost [45]. The versatility of ATLO-ML was further demonstrated through validation beyond traditional machine learning approaches by incorporating Fujitsu's open source AutoML technology, SapientML [46, 47, 48, 49]. Results indicate that ATLO-ML significantly improves prediction accuracy while providing users with a user-friendly interface to adjust output length and obtain optimized input time length and sampling rate parameters. This approach not only enhances predictive performance but also offers a more adaptable and efficient solution for time series analysis across various domains.

The remainder of the paper is structured as follows. Section II elucidates the data collection methodologies and data pre-processing, encompassing both public datasets and the GAMS-dataset. Section II presents a detailed explanation of the validation data collected within the Data Center, including comprehensive information on the sensors employed, their technical specifications, the duration of data collection, and the various types of sensory data acquired. Section III expounds upon the application of ATLO-ML for processing time series signals. Section III provides a thorough exposition of the complete ATLO-ML procedure, detailing each step of the process. Section IV demonstrates the comparative efficacy of various machine learning methodologies before and after their integration with ATLO-ML. The analysis encompasses a range of datasets and algorithmic validations, followed by an in-depth discussion of the experimental outcomes. The paper concludes with Section V, which synthesizes the key findings of the research. Section V articulates the specific challenges addressed by ATLO-ML and highlights its distinct advantages in the domain of time series analysis and machine learning optimization. In the final of the paper, Section VI discussed the feature work.

## II. DATASET COLLECTION & PRE-PROCESSING

This study leverages both the publicly available GAMS-dataset and a proprietary dataset for comprehensive validation. The subsequent section offers a detailed exposition of the utilized data, encompassing the specifications of employed sensors, the parameters measured in the data collection process, and the pre-processing methodologies applied. This dual-dataset approach enables a robust validation framework for the research findings.

### A. GAMS Indoor Air Quality Dataset

This study utilizes the GAMS-dataset as a public dataset, which comprises indoor and outdoor air quality data collected by GAMS Environmental Monitoring [50], a Chinese company specializing in real-time air quality monitoring and airborne viral risk assessment for indoor environments. The analysis focuses specifically on the indoor data collected by sensors deployed in Shanghai, China. The indoor dataset encompasses six key parameters: $CO_2$, humidity, $PM_{10}$, $PM_{2.5}$, temperature, and VOCs. The temporal scope of the dataset spans from November 2016 to March 2017, encompassing a total of 135,099 data points. The data was aggregated into hourly intervals, resulting in approximately 3,000 observations.

### B. Data Center Air Quality Dataset

For the proprietary dataset collection, Internet of Things (IoT) technology was employed, an approach consistent with previous air quality monitoring studies [28]. The infrastructure utilizes the A150 system from Archimedes Controls Corp., which transmits data to ARCOS (IoT Edge Server) and subsequently to Fujitsu Server for computational analysis. The A150 environmental sensor system is engineered for diverse IoT applications, including data centers, IT infrastructure, industrial control, automation, agriculture, food safety, transportation, environmental and building management. It offers capabilities for environmental and physical security, energy efficiency, long-term transparency, visibility, and remote management. Fig. 1 illustrates the detailed data collection architecture. The A150 system efficiently gathered comprehensive data center metrics, including $PM_1$, $PM_{10}$, $P_{0.3}$, $P_{0.5}$, $P_{1.0}$, humidity, temperature, pressure, and air quality index (AQI). The sensor installation diagram in the data center server is shown in Fig. 2.

Data was collected continuously (24/7) from July 2024 to October 2024. The sampling rates for the sensors were as follows: air particle sensor every two minutes, air pressure sensor every second, and both temperature and humidity sensors every 30 seconds.

*C. Data Pre-processing*

This study employed a comprehensive data pre-processing pipeline to ensure the quality and consistency of sensor data for subsequent analysis. The process encompassed several key steps:

1) Data Acquisition

   Sensor readings are retrieved from a structured database within a specified time frame $[t_{start}, t_{end}]$, where $t_{end}$ represents the most recent timestamp, and $t_{start}$ is defined as:

   $$t_{start} = t_{end} - \Delta t, \quad (1)$$

   where $\Delta t$ is the desired time range for analysis, typically set to the most recent n days.

2) Temporal Alignment and Resampling

   A temporal alignment process is implemented to standardize data across sensors with varying sampling rates. Given a set of timestamps $T = \{t_0, t_1, \ldots, t_n\}$ and a desired sampling interval R, a new time series $\dot{T}$ is create such that:

   $$\dot{T} = \{\dot{t} \mid \dot{t} = t_1 + k \times R, k \in \mathbb{N}, \dot{t} \leq t_n\}. \quad (2)$$

   For each sensor $S_j$, then its readings then be aligned to $\dot{T}$ using the following rule:

   $$S_j(\dot{t}) = S_j(t_i), \text{where } t_i = \max\{t \in T \mid t \leq \dot{t}\}. \quad (3)$$

3) Handling Missing Data

   To address the missing data, the Forward Fill method is exclusively employed. For a time series $X = \{x_1, x_2, \ldots, x_n\}$ with missing values, the Forward Fill operation $FF(X)$ is defined as:

   $$FF(X)(t) = x_t \text{ if } x_{t_{prev}} \in x_t, \quad (4)$$

   otherwise,

   $$t_{prev} = \max\{\tau < t \mid x_\tau \neq \emptyset\}. \quad (5)$$

   This method propagates the last valid observation forward to fill gaps, which is particularly suitable for sensors with relatively stable readings over short periods.

4) Constant Column Elimination

   For each sensor $S_j$, its variance $\sigma_j^2$ is computed. If $\sigma_j^2 < \varepsilon$, where $\varepsilon$ is a predefined threshold close to zero, $S_j$ will be removed from the dataset.

5) Relevant Measurement Selection

   Let R be the set of relevant measurement types, and $\mu(S_j)$ be the measurement type of sensor $S_j$. Sensor set $\dot{S}_j$ is filtered out as follows:

   $$\dot{S}_j = \{S_j \in S \mid \mu(S_j) \in R\}. \quad (6)$$

6) Data Restructuring

   The data from is transformed into a long format (timestamp, $S_j$, value) to a wide format matrix D where:

   $$D[i, j] = S_j(t_i). \quad (7)$$

   This restructuring facilitates time series analysis and model input preparation.

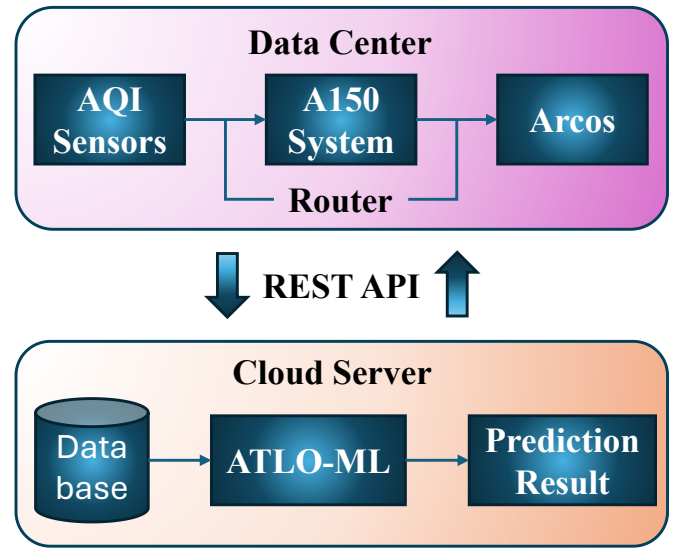

Fig. 1. Data collection structure for data center.

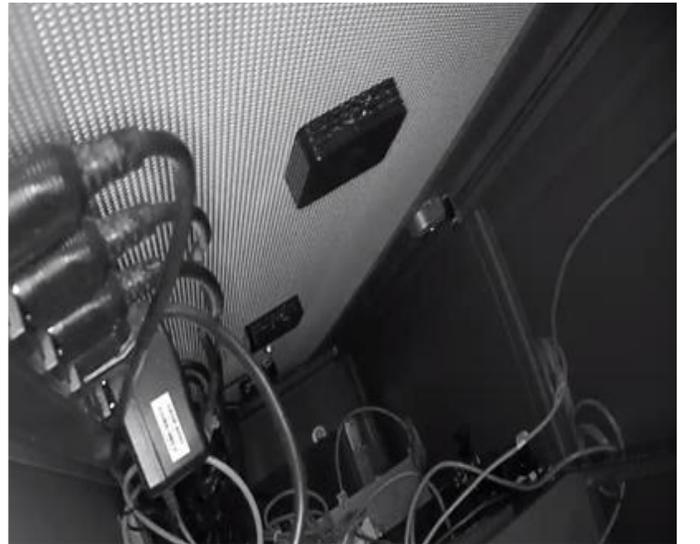

Fig. 2. The sensor installation in the data center server.

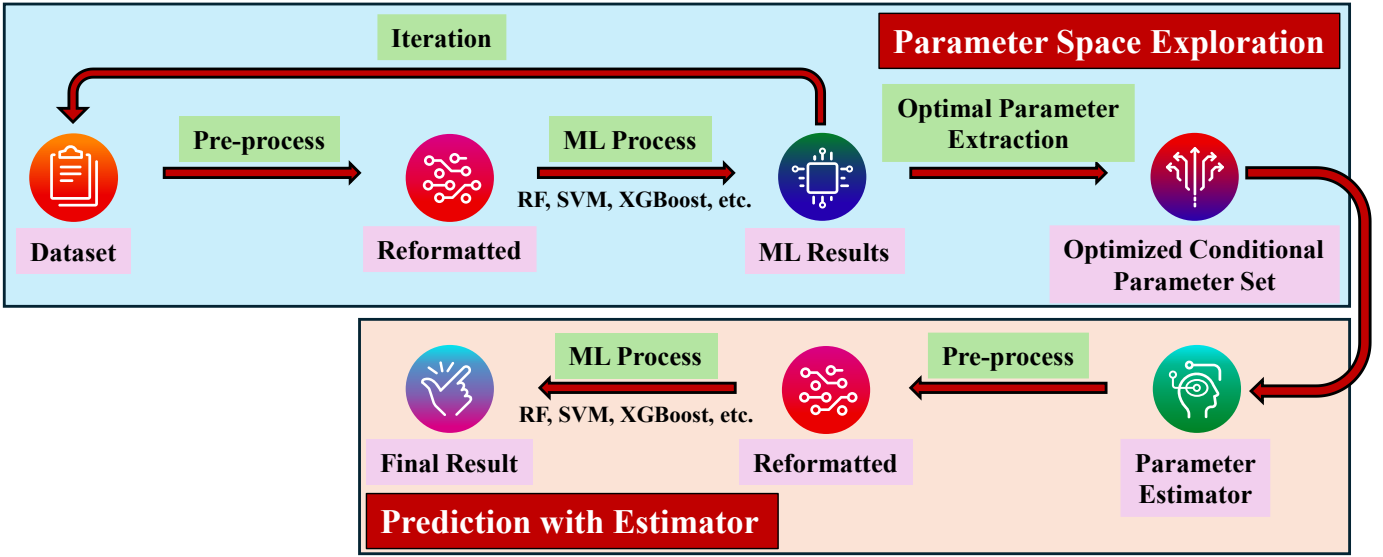

Fig. 3. The flow chart diagram of ATLO-ML.

7) Time Window Creation

For predictive modeling, input-output pairs (X, Y) is generated where:

$$X = D[t - I/R: t, :], \tag{8}$$

represents a window of I/R time steps up to time t, I represents the set of input time windows, R represents the set of data sampling rates.

$$Y = D[t + 1: t + H/R, :], \tag{9}$$

represents the H/R time steps immediately following t, H represents the set of output time horizons. This window slides across the entire dataset to create a comprehensive set of training examples.

Through this systematic pre-processing methodology, raw sensor data is transformed into a structured, normalized, and analysis-ready format, setting a solid foundation for subsequent modeling and analytical tasks. It's important to note that the sampling interval R, input window size I, and output window size H used in the time window creation are determined through the ATLO-ML. ATLO-ML takes a specified time period as input and helps us identify the optimal time scale for prediction within that period. This approach ensures that the input windows capture the most relevant temporal dynamics for the forecasting tasks. By adapting the time scale to the specific prediction period of interest, the approach enhances the robustness and accuracy of subsequent modeling efforts, allowing for more precise and reliable forecasts across various time horizons.

## III. ATLO-ML

This section introduces ATLO-ML, a framework comprising two primary components: Parameter Space Exploration and Prediction with Estimator. The Parameter Space Exploration phase facilitates the acquisition of a substantial set of optimized parameter configurations within the parameter space. Subsequently, the Prediction with Estimator component utilizes an estimator in conjunction with the obtained parameter configurations to estimate optimized parameters, which are then employed in the execution of machine learning algorithms to get the optimized result. In this case, the parameters are input length, output length, and sample rate. The diagram of ATLO-ML is shown in Fig. 3.

### A. Parameter Space Exploration

Parameter Space Exploration in the ATLO-ML framework is a systematic approach to investigating the multidimensional hyperparameter space of machine learning models. This process is fundamental to optimizing model performance across various data configurations and prediction tasks.

Let $\Theta$ denote the hyperparameter space, which is defined by the Cartesian product of individual parameter sets:

$$\Theta = H \times R \times I, \tag{10}$$

where $H = \{h_1, h_2, \ldots, h_n\}$ represents the set of output time horizons; $R = \{r_1, r_2, \ldots, r_m\}$ represents the set of data sampling rates; $I = \{i_1, i_2, \ldots, i_k\}$ represents the set of input time windows.

For each configuration $\theta \in \Theta$, where $\theta = (h, r, i)$, a machine learning model $M_\theta$ including pre-processing, training and evaluation. The pre-processing is described in the previous section. The evaluation process can be formalized as:

$$E(M_\theta) = w_1 \times R^2(M_\theta) + (1 - w_1) \times [w_2 \times \text{NRMSE}(M_\theta) + w_3 \times \text{NRMSLE}(M_\theta) + w_4 \times \text{NMAE}(M_\theta)], \tag{11}$$

where $E(M_\theta)$ is the overall evaluation score,

$$w_1 + w_2 + w_3 + w_4 = 1 \tag{12}$$

and NRMSE, NRMSLE, and NMAE are normalized versions of Root Mean Square Error (RMSE), Root Mean Square Logarithmic Error (RMSLE), and Mean Absolute Error (MAE) respectively. The normalization process for each error metric could be:

$$\text{NRMSE} = 1 - \frac{\text{RMSE}}{\max(\text{RMSE})},$$

$$\text{NRMSLE} = 1 - \frac{\text{RMSLE}}{\max(\text{RMSLE})}, \quad (13)$$

$$\text{NMAE} = 1 - \frac{\text{MAE}}{\max(\text{MAE})},$$

where $\max(\text{RMSE})$, $\max(\text{RMSLE})$, and $\max(\text{MAE})$ are the maximum values of these metrics across all models in $\Theta$. Coefficient of determination ($R^2$) is shown as follows:

$$R^2 = 1 - \frac{\sum(y_i - \hat{y}_i)^2}{\sum(y_i - \bar{y}_i)^2}, \quad (14)$$

where $y_i$ are the observed values, $\hat{y}_i$ are the predicted values, and $\bar{y}_i$ is the mean of observed values. $R^2 \in [0, 1]$, with 1 indicating perfect prediction. RMSE is shown as follows:

$$\text{RMSE} = \sqrt{\sum_{i=1}^{n} \frac{(\hat{y}_i - y_i)^2}{n}}, \quad (15)$$

where n is the number of observations. RMSE provides the standard deviation of prediction errors. RMSLE is shown as follows:

$$\text{RMSLE} = \sqrt{\frac{\sum_{i=1}^{n}(\log(y_i+1) - \log(\hat{y}_i+1))^2}{n}}, \quad (16)$$

RMSLE is particularly useful for datasets with exponential growth and when the goal is to penalize underestimates more than overestimates. MAE is shown as follows:

$$\text{MSE} = \sum_{i=1}^{n} \frac{|y_i - \hat{y}_i|}{n}, \quad (17)$$

MAE provides the average magnitude of errors in prediction, without considering their direction. By normalizing RMSE, RMSLE, and MAE, the normalization process ensures all metrics are on a 0-1 scale, with higher values indicating better performance. The overall score will be between 0 and 1, with 1 being the best possible score. The weights can be adjusted based on the specific requirements of the prediction task. $R^2$ is separated from the error metrics, allowing for easy adjustment of the balance between goodness-of-fit ($R^2$) and prediction error measures. By including multiple normalized error metrics, the evaluation framework captures different aspects of model performance without any single metric dominating. This approach provides a balanced, interpretable, and flexible evaluation function that can be easily adapted to different scenarios within the ATLO-ML framework.

The exploration process generates a set of evaluated models:

$$\Omega = \{(\theta, E(M_\theta)) \mid \theta \in \Theta\}. \quad (18)$$

Then the process proceeds to find the best parameters for each output time horizon $h \in H$. This step can be formalized as:

$$\dot{\theta}_h = \arg\max\{E(M_\theta): \theta \in \Theta_h\}, \quad (19)$$

$$\Theta_h = \{\theta \in \Theta \mid \theta = (h, r, i) \text{ for } h \in H, i \in I\}. \quad (20)$$

This process yields a set of optimal parameters for each output time horizon:

$$\Phi = \{(h, \dot{\theta}_h) \mid h \in H\}. \quad (21)$$

The set $\Phi$ represents the optimal conditional parameter set, which forms the basis for the subsequent Prediction with Estimator phase, where H is the set of explored output time horizons and $\dot{\theta}_h$ represents the optimal parameter configuration for time horizon h.

The Parameter Space Exploration phase is crucial as it provides:
- A comprehensive mapping of the parameter space to model performance.
- Identification of optimal parameter configurations for each output time horizon.
- A foundation for estimating parameters for unseen output time horizons.

This thorough exploration and optimization process enhances the robustness and adaptability of the machine learning models in the context of time series prediction and analysis. It enables the framework to make informed decisions about hyperparameter selection for various prediction tasks and data characteristics, ultimately leading to more accurate and reliable predictions.

*B. Prediction with Estimator*

The Prediction with Estimator component of ATLO-ML employs a sophisticated approach to parameter estimation for unseen output time horizons. For a new, unseen output time horizon $\dot{h} \notin H$, the estimator E aims to approximate the optimal parameter configuration $\dot{\theta}_{\dot{h}}$:

$$\dot{\theta}_{\dot{h}} = E(\dot{h}, \Phi). \quad (22)$$

The estimator E is implemented through various regression techniques, each offering distinct advantages:

1) Linear Interpolation/Extrapolation (LI/E):
   This method assumes a linear relationship between output time horizons and optimal parameters. For $\dot{h} \in [\min(H), \max(H)]$, linear interpolation is used:

   $$\dot{\theta}_{\dot{h}} = \theta_{h_1} + (\dot{h} - h_1) \times \frac{\theta_{h_2} - \theta_{h_1}}{h_2 - h_1}, \quad (23)$$

   where $h_1, h_2 \in H$ are the closest known time horizons to $\dot{h}$. For $\dot{h} \notin [\min H, \max H]$, linear extrapolation is employed.

2) Polynomial Regression (PR):
   This approach models the relationship between h and $\dot{\theta}_{h_1}$ as a polynomial of degree d:

   $$\dot{\theta}_{\dot{h}} = \sum_{k=0}^{d} a_k \times \dot{h}^k, \quad (24)$$

   where coefficients $a_i$ are estimated using least squares fitting on $\Phi$.

3) Exponential Smoothing (ES):
   This method applies weighted averages, giving more importance to recent observations:

   $$\dot{\theta}_{\dot{h}} = \alpha \times S_h + (1 - \alpha) \times S_{h_{-1}}, \quad (25)$$

   where $\alpha \in (0, 1)$ is the smoothing factor and $S_h$ is the smoothed statistic.

4) KNN Regression:
   This non-parametric approach estimates $\dot{\theta}_{\dot{h}}$ based on

the k nearest neighbors in $\Phi$:

$$\Phi: \dot{\theta}_{\dot{h}} = \frac{1}{k} \times \sum_{i=0}^{k} \theta_{h_i}, \quad (26)$$

where $o_i$ are the k closest time horizons to $\dot{h}$ in H.

The estimator employs these techniques to approximate both the input length and sampling rate for the unseen output time horizon. To ensure practical viability, constraints are applied:

- The estimated input length i is bounded by the output length:

$$i = \max(\dot{\theta}_{\dot{h}}[i], \dot{h}), \quad (27)$$

- The estimated sampling rate $\dot{r}$ is subject to a dynamic lower bound:

$$\dot{r} = \max(\dot{\theta}_{\dot{h}}[\dot{r}], \beta) \quad (28)$$

where $\beta$ ensures a minimum number of samples per output length.

This multi-faceted approach to parameter estimation allows the ATLO-ML framework to adapt to a wide range of output time horizons, enhancing its flexibility and applicability across diverse time series prediction tasks.

## IV. EXPERIMENTAL RESULTS

This section presents the experimental results of applying ATLO-ML to predict time series air quality datasets. The analysis examined two datasets: the GAMS-dataset and a proprietary collection of data center air quality measurements. The analysis focuses on machine learning predictions derived from parameters suggested by the ATLO-ML estimator. The findings demonstrate the efficacy of this approach in forecasting air quality trends across diverse environmental contexts.

In parameter space exploration, H = {5, 10, 2, 40, 80, 120, 160, 200, 240, 280, 320, 260, 400, 440, 480, 520, 260, 600, 640} min, R = H/[1, 2, 3, 4], I = H × [1, 2, 3, 4]. In terms of validation output time horizon $\dot{H}$ = {8, 15, 30, 60, 100, 140, 180, 220, 260, 300, 340, 380, 420, 460, 500, 580, 620} min. If without ATLO-ML Estimator, R = H/3 and I = H × 3.

The experimental results are presented in TABLE I and II, where cells shaded in yellow indicate the best performance within each machine learning model for a given feature, and cells shaded in purple denote the best overall performance across all machine learning methods for a specific feature. Consistently, the ATLO-ML estimator demonstrates superior performance compared to fixed input length and sample rate approaches.

TABLE I
THE EXPERIMENTAL RESULTS ($R^2$) OF GAMS-DATASET WITH AND WITHOUT ATLO-ML

| ML Method | Estimator | Humi | Temp | $PM_{2.5}$ | $CO_2$ | $PM_{10}$ | VOCs |
|---|---|---|---|---|---|---|---|
| DT | LI/E | 0.8921 | 0.8386 | 0.8961 | 0.8071 | 0.8903 | 0.4540 |
| | PR | 0.8911 | 0.8225 | 0.8868 | 0.8017 | 0.8774 | 0.4405 |
| | ES | 0.8988 | 0.8420 | 0.8908 | 0.8078 | 0.8894 | 0.4296 |
| | KNN | 0.8904 | 0.8290 | 0.8929 | 0.8023 | 0.8804 | 0.4477 |
| | None | 0.8534 | 0.7215 | 0.9828 | 0.6950 | 0.9821 | -0.0345 |
| KNN | LI/E | 0.9136 | 0.8955 | 0.9131 | 0.8911 | 0.9081 | 0.7125 |
| | PR | 0.9179 | 0.8930 | 0.9132 | 0.8888 | 0.9078 | 0.7038 |
| | ES | 0.9139 | 0.8945 | 0.9131 | 0.8983 | 0.9119 | 0.7115 |
| | KNN | 0.9152 | 0.8919 | 0.9131 | 0.8898 | 0.9100 | 0.7083 |
| | None | 0.7934 | 0.7246 | 0.7626 | 0.6968 | 0.7672 | 0.4475 |
| LightGBM | LI/E | 0.9707 | **0.9555** | **0.9951** | 0.9502 | 0.9940 | 0.7354 |
| | PR | 0.9707 | 0.9552 | 0.9950 | 0.9504 | 0.9939 | 0.7352 |
| | ES | 0.9696 | 0.9547 | 0.9948 | 0.9506 | 0.9939 | 0.7257 |
| | KNN | **0.9714** | 0.9537 | 0.9950 | 0.9497 | 0.9943 | 0.7369 |
| | None | 0.9286 | 0.8794 | 0.9801 | 0.8783 | 0.9802 | 0.5292 |
| RF | LI/E | 0.9540 | 0.9178 | 0.9644 | 0.9221 | 0.9617 | 0.6888 |
| | PR | 0.9489 | 0.9151 | 0.9644 | 0.9133 | 0.9618 | 0.6781 |
| | ES | 0.9534 | 0.9195 | 0.9644 | 0.9242 | 0.9621 | 0.6862 |
| | KNN | 0.9511 | 0.9151 | 0.9648 | 0.9162 | 0.9621 | 0.6849 |
| | None | 0.9247 | 0.8689 | 0.9911 | 0.8585 | 0.9905 | 0.5346 |
| SVM | LI/E | 0.9699 | 0.9410 | 0.9949 | 0.9233 | 0.9947 | 0.6047 |
| | PR | 0.9711 | 0.9334 | 0.9946 | 0.9134 | **0.9950** | 0.6046 |
| | ES | **0.9713** | 0.9405 | 0.9950 | 0.9219 | 0.9946 | 0.6121 |
| | KNN | 0.9695 | 0.9323 | 0.9948 | 0.9146 | 0.9949 | 0.6014 |
| | None | 0.8983 | 0.8516 | 0.9414 | 0.8319 | 0.9443 | 0.5226 |
| XGBoost | LI/E | 0.9664 | 0.9462 | 0.9951 | 0.9422 | 0.9938 | 0.7122 |
| | PR | 0.9674 | 0.9473 | 0.9950 | 0.9426 | 0.9942 | 0.7185 |
| | ES | 0.9668 | 0.9464 | 0.9949 | 0.9423 | 0.9940 | 0.7052 |
| | KNN | 0.9672 | 0.9467 | 0.9950 | 0.9410 | 0.9942 | 0.7193 |
| | None | 0.9194 | 0.8623 | 0.9907 | 0.8513 | 0.9899 | 0.4517 |
| AutoML (SapientML) | LI/E | 0.9705 | 0.9528 | 0.9951 | 0.9492 | 0.9936 | 0.7434 |
| | PR | 0.9694 | 0.9521 | 0.9949 | 0.9520 | 0.9936 | **0.7489** |
| | ES | 0.9697 | 0.9514 | 0.9945 | **0.9522** | 0.9940 | 0.7442 |
| | KNN | 0.9701 | 0.9526 | 0.9949 | 0.9505 | 0.9935 | 0.7463 |
| | None | 0.9490 | 0.9085 | 0.9946 | 0.8877 | 0.9941 | 0.5756 |

Note: Humi = Humidity, Temp = Temperature

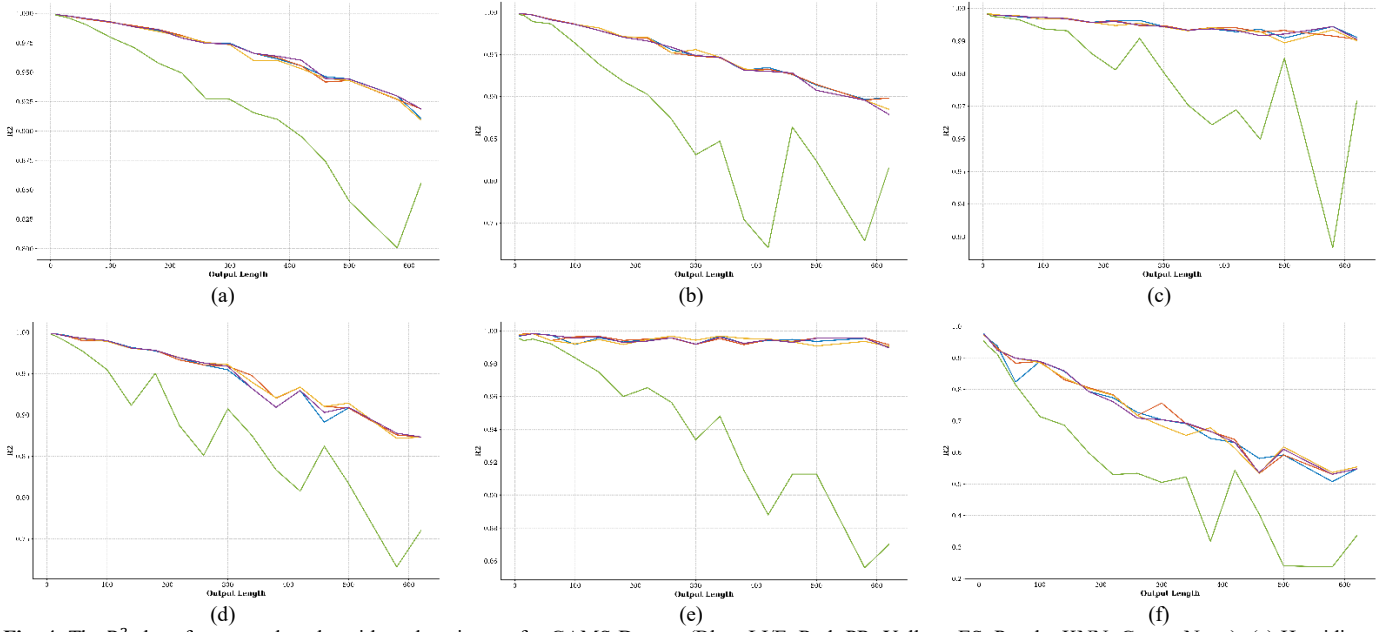

**Fig. 4.** The $R^2$ chart for output lengths with each estimator for GAMS-Dataset (Blue: LI/E, Red: PR, Yellow: ES, Purple: KNN, Green: None): (a) Humidity – LightGBM; (b) Temperature – LightGBM; (c) $PM_{25}$ – LightGBM; (d) $CO_2$ – AutoML; (e) $PM_{10}$ –SVM; (f) VOCs – AutoML.

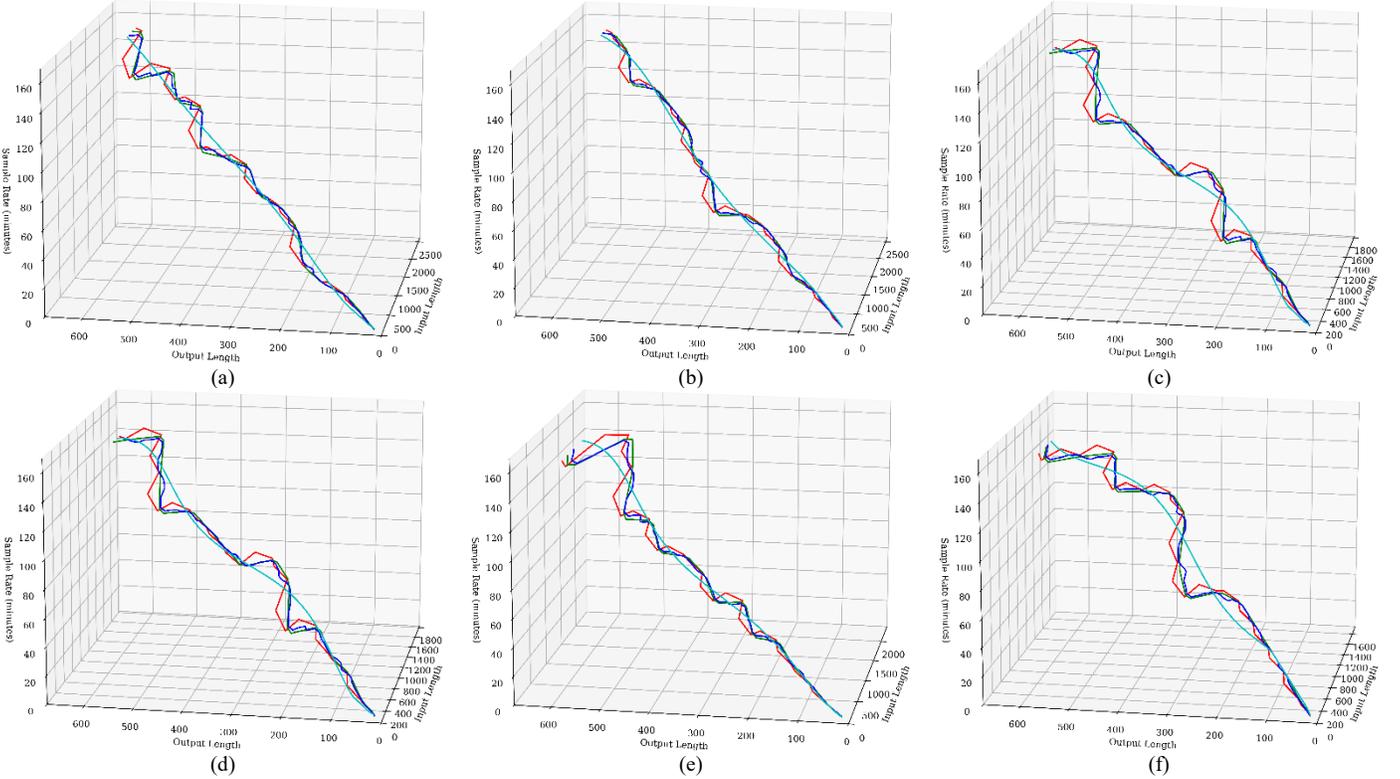

**Fig. 5.** 3D parameter space visualization (Red: ES, Green: LI/E, Blue: KNN, Cerulean: PR): (a) Humidity – LightGBM; (b) Temperature – LightGBM; (c) $PM_{25}$ – LightGBM; (d) $CO_2$ – AutoML; (e) $PM_{10}$ –SVM; (f) VOCs – AutoML.

*A. Results of GAMS-Dataset*

TABLE I displays the experimental results, reported as $R^2$ values, for different machine learning methods applied to the GAMS-Dataset. The methods compared include DT, KNN, LightGBM, RF, SVM, XGBoost, and AutoML. For each method, several estimators are evaluated: L1/E, PR, ES, KNN, and a baseline without any estimator (None).

In the GAMS-Dataset experiment, among the machine learning methods evaluated, LightGBM with KNN demonstrated the highest accuracy in predicting humidity, while the LI/E estimator achieved the highest accuracy for temperature and $PM_{25}$ predictions. For humidity and $PM_{10}$ predictions, the SVM with ES estimator and the PR estimator were the most accurate, respectively. XGBoost, combined with the LI/E estimator, also excelled in predicting $PM_{25}$. Additionally, AutoML, paired with the LI/E estimator, ES estimator, and PR, demonstrated the highest accuracy for predicting $PM_{25}$, $CO_2$, and VOC, respectively.

Interestingly, when examining AutoML without an estimator, its performance surpasses all other machine learning methods evaluated. However, with the proposed estimator, even a simple ATLO-ML model outperforms AutoML. This observation underscores the effectiveness of the approach in handling time series datasets, demonstrating that a well-designed estimator can enhance the performance of simpler models to surpass more complex AutoML solutions.

The varied performance across different environmental parameters indicates that some factors (e.g., $PM_{2.5}$, $PM_{10}$) are more predictable than others (e.g., VOCs) using these methods. This could be due to the inherent complexity of the underlying processes or the quality and relevance of the available features for each parameter. Future research directions could involve exploring why certain methods excel for specific parameters and investigating ways to improve performance on the more challenging parameters like VOCs.

As shown in Fig. 4, this study presents an $R^2$ chart depicting the performance of various estimators across different output lengths. To maintain conciseness, the analysis exclusively shows the ATLO-ML models that demonstrate the highest value for each feature. Specifically, the $R^2$ values are displayed for LightGBM predicting humidity, temperature, and $PM_{2.5}$; AutoML predicting $CO_2$ and VOCs; and SVM predicting $PM_{10}$. The chart employs color-coded lines to represent the $R^2$ values of different estimators: blue for Linear Interpolation/Extrapolation, orange for PR, yellow for ES, purple for KNN Regression, and green for scenarios without an estimator. Each line illustrates the performance trajectory of its respective estimator across the spectrum of output lengths examined in this analysis.

As illustrated in Fig. 4, there is a discernible trend of decreasing $R^2$ values as the prediction time horizon extends, indicating a progressive decline in predictive accuracy over longer temporal intervals. Across all the features examined, the results demonstrate that in the absence of ATLO-ML estimators, the deterioration in predictive accuracy becomes markedly pronounced with increasing time spans. Conversely, the implementation of ATLO-ML estimators exhibits a significant capacity to mitigate the accuracy degradation associated with extended time horizons. A particularly striking observation is that, irrespective of the chosen prediction interval, the performance of non-ATLO-ML approaches consistently fails to surpass that of ATLO-ML estimators, underscoring the latter's superior predictive capabilities across varying temporal scales.

Fig. 5 illustrates the parameter spaces derived from various estimators. The red line, green line, blue line, and cerulean line show the parameter space of ES estimator, LI/E estimator, KNN estimator, and PR estimator, respectively. To maintain conciseness, only the estimators with the highest $R^2$ values are presented. Specifically, LightGBM was utilized for predicting Humidity, Temperature, and $PM_{2.5}$; AutoML for $CO_2$ and VOCs; and SVM for $PM_{10}$. While numerous studies arbitrarily set the input length to thrice the output length and fix the sample rate to match the output length, this investigation reveals that such an approach does not yield optimal results. The parameter space diagrams demonstrate that different predictive features require distinct ratios of input length and sample rate. Some features necessitate input lengths of 2 to 3 times the output length, while others demand 3 to 4 times. Regarding the sample rate in the GAMs-dataset, it exhibits remarkable consistency across features, ranging between $\frac{1}{4}$ to $\frac{1}{3}$ of the output length.

The GAMS-Dataset experiment presents results comparing various machine learning models and estimators for predicting environmental parameters. The ATLO-ML estimator consistently outperformed fixed input length and sample rate approaches across different features. Notably, even simple ATLO-ML models surpassed the performance of more complex AutoML solutions like SapientML when equipped with the proposed estimator. Analysis of $R^2$ values across different output lengths revealed that ATLO-ML estimators effectively mitigated accuracy degradation over extended time horizons compared to non-ATLO-ML approaches. Furthermore, the investigation of parameter spaces demonstrated that optimal input length and sample rate ratios vary among predictive features, challenging the conventional practice of using fixed ratios.

*B. Results of Our Data Center Measurement Dataset*

TABLE II displays the experimental results, reported as $R^2$ values, for different machine learning methods applied to the Dataset that is collected within the data center. The methods compared include DT, KNN, LightGBM, RF, SVM, XGBoost, and AutoML. For each method, several estimators are evaluated: LI/E, PR, ES, KNN, and a baseline without any estimator (None). Comparing Tables I and II reveals that the dataset poses greater challenges for machine learning models. Using identical hyperparameters, the results in Table II demonstrate poorer performance compared to Table I.

Within the experimental investigation, the comparative analysis of various machine learning algorithms revealed that the LightGBM incorporating LI/E estimator exhibited superior predictive accuracy. KNN estimator demonstrated optimal performance in humidity and pressure predictions, while the PR estimator yielded the highest accuracy metrics in temperature forecasting. The analysis revealed that the XGBoost achieved optimal $R^2$ values across multiple air quality parameters when integrated with specific estimators. Specifically, XGBoost with PR estimator demonstrated superior predictive performance for $PM_{10}$, ES estimator for $P_{0.3}$, PR estimator for $P_{0.5}$, KNN estimator for $P_{1.0}$, and PR estimator maximized AQI prediction accuracy.

The empirical findings demonstrate that estimators generally improve predictive accuracy when implemented with properly functioning models. However, this enhancement effect is notably absent in cases where the base models exhibit inherent performance limitations. This phenomenon was particularly evident with SVM and AutoML, where the integration of estimators failed to yield significant improvements in model performance.

In the comparative analysis of machine learning frameworks, an unexpected finding emerged regarding the performance of AutoML on data center proprietary dataset. Despite its reputation for automated machine learning excellence, AutoML exhibited substantially suboptimal predictive performance. Through comprehensive investigation of its automated preprocessing pipeline, the analysis identified a critical limitation in its data handling methodology that significantly impaired model training efficacy. The primary issue resides in AutoML's automated preprocessing pipeline, specifically its uniform random sampling approach. The framework automatically reduces the dataset to 100,000 instances across twelve target variables, comprising three distinct air measurement parameters and time horizons (t + 1 through t + n). This sampling methodology introduces several significant challenges:

- Disruption of Temporal Coherence: The random sampling mechanism fundamentally compromises the temporal continuity essential for time series prediction, thereby degrading the model's capacity to capture sequential patterns and temporal dependencies.
- Insufficient Resolution for Multi-target Complexity: The dimensionality of twelve concurrent target variables, representing various parameters and time horizons, demands a larger sample size to adequately maintain the intricate relationships between interconnected measurements.
- Critical Information Loss: The fixed sample size of 100,000 instances proves inadequate in representing the full complexity and variability inherent in the original dataset, particularly regarding rare but significant events or patterns.
- Degradation of Cross-variable Dependencies: The uniform sampling approach fails to preserve crucial correlations between different air measurement parameters, leading to suboptimal model learning and reduced predictive accuracy.

While the implementation of estimators demonstrated significant potential in enhancing machine learning prediction accuracy, the findings reveal an important limitation in their application. Notably, when analyzing $P_{0.3}$ predictions using AutoML, the integration of ATLO-ML estimators markedly improved the $R^2$ from negative values to exceeding 0.7. However, this enhancement mechanism proved ineffective when the underlying machine learning process itself was fundamentally flawed. This phenomenon was particularly evident in the prediction of other air quality indices, where $R^2$ values remained negative despite estimator implementation. These results suggest that while estimators can substantially augment the performance of well-structured machine learning processes, they cannot compensate for or correct fundamental deficiencies in the base algorithmic approach. This underscores the critical importance of ensuring the validity of the core machine learning methodology before applying performance-enhancing estimators.

TABLE II
THE EXPERIMENTAL RESULTS ($R^2$) OF DATA CENTER DATASET WITH AND WITHOUT ATLO-ML

| ML Method | Estimator | $PM_1$ | $PM_{10}$ | $P_{0.3}$ | $P_{0.5}$ | $P_{1.0}$ | Humi | Temp | Pres | AQI |
|---|---|---|---|---|---|---|---|---|---|---|
| DT | LI/E | 0.8523 | 0.7996 | 0.7486 | 0.8519 | 0.8170 | 0.6318 | 0.5426 | 0.7444 | 0.8510 |
|  | PR | 0.8601 | 0.7900 | 0.7585 | 0.8704 | 0.8088 | 0.6408 | 0.5371 | 0.7244 | 0.8323 |
|  | ES | 0.8486 | 0.7749 | 0.7414 | 0.8643 | 0.8278 | 0.6147 | 0.5341 | 0.7196 | 0.8416 |
|  | KNN | 0.8754 | 0.7723 | 0.7467 | 0.8492 | 0.9174 | 0.6417 | 0.5359 | 0.7367 | 0.8331 |
|  | None | 0.7615 | 0.7278 | 0.5822 | 0.7601 | 0.6896 | 0.3431 | 0.1374 | 0.3070 | 0.7810 |
| KNN | LI/E | 0.8272 | 0.7734 | 0.7635 | 0.8344 | 0.8016 | 0.7393 | 0.6936 | 0.7938 | 0.7990 |
|  | PR | 0.8202 | 0.7729 | 0.7568 | 0.8267 | 0.7986 | 0.7438 | 0.6923 | 0.7770 | 0.8014 |
|  | ES | 0.7965 | 0.7628 | 0.7574 | 0.8135 | 0.7935 | 0.7402 | 0.6892 | 0.7974 | 0.7965 |
|  | KNN | 0.8234 | 0.7717 | 0.7692 | 0.8357 | 0.8028 | 0.7521 | 0.6913 | 0.7910 | 0.8036 |
|  | None | 0.6447 | 0.6101 | 0.5468 | 0.6424 | 0.6279 | 0.5456 | 0.4494 | 0.4976 | 0.6377 |
| LightGBM | LI/E | 0.9365 | 0.8826 | 0.8998 | 0.9432 | 0.9064 | 0.8499 | 0.7989 | 0.9342 | 0.9053 |
|  | PR | 0.9323 | 0.8896 | 0.8990 | 0.9418 | 0.9048 | 0.8563 | **0.8043** | 0.9350 | 0.9051 |
|  | ES | 0.9268 | 0.8961 | 0.8969 | 0.9393 | 0.9026 | 0.8571 | 0.8027 | 0.9330 | 0.9098 |
|  | KNN | 0.9267 | 0.8859 | 0.9019 | 0.9440 | 0.9132 | **0.8585** | 0.8019 | **0.9358** | 0.9068 |
|  | None | 0.8548 | 0.8177 | 0.7936 | 0.8592 | 0.8230 | 0.6987 | 0.5214 | 0.7224 | 0.8399 |
| RF | LI/E | 0.9331 | 0.8967 | 0.8686 | 0.9400 | 0.9044 | 0.8085 | 0.7520 | 0.8792 | 0.9188 |
|  | PR | 0.9353 | 0.8862 | 0.8704 | 0.9367 | 0.8999 | 0.8131 | 0.7581 | 0.8701 | 0.9130 |
|  | ES | 0.9197 | 0.8897 | 0.8624 | 0.9294 | 0.9048 | 0.8139 | 0.7546 | 0.8810 | 0.9060 |
|  | KNN | 0.9308 | 0.8864 | 0.8642 | 0.9400 | 0.9066 | 0.8096 | 0.7513 | 0.8778 | 0.9117 |
|  | None | 0.8935 | 0.8664 | 0.7829 | 0.8990 | 0.8672 | 0.6399 | 0.5307 | 0.5825 | 0.8946 |
| SVM | LI/E | -5.9877 | -5.0284 | -27.593 | -1.9355 | -10.292 | -64.180 | -77.978 | -18.109 | -4.6989 |
|  | PR | -5.8046 | -23.108 | -25.648 | -1.4980 | -7.7875 | -88.487 | -62.832 | -18.380 | -7.2501 |
|  | ES | -3.8811 | -11.755 | -22.884 | -1.5586 | -9.7168 | -61.261 | -119.76 | -18.892 | -6.0020 |
|  | KNN | -7.0714 | -22.836 | -33.237 | -1.4937 | -10.657 | -82.935 | -117.32 | -20.596 | -15.915 |
|  | None | -0.3533 | -0.1433 | -15.222 | -0.3576 | -3.0697 | -72.394 | -35.410 | -15.123 | -0.3118 |
| XGBoost | LI/E | 0.9567 | 0.9147 | 0.9209 | 0.9686 | 0.9396 | 0.8282 | 0.7689 | 0.9211 | 0.9376 |
|  | PR | 0.9563 | 0.9298 | 0.9217 | **0.9700** | 0.9365 | 0.8275 | 0.7657 | 0.9200 | **0.9449** |
|  | ES | 0.9571 | 0.9032 | **0.9259** | 0.9676 | 0.9400 | 0.8267 | 0.7670 | 0.9202 | 0.9290 |
|  | KNN | **0.9639** | **0.9168** | 0.9220 | 0.9695 | **0.9417** | 0.8225 | 0.7714 | 0.9234 | 0.9408 |
|  | None | 0.8894 | 0.8477 | 0.8182 | 0.9097 | 0.8500 | 0.6599 | 0.4541 | 0.7169 | 0.8987 |
| AutoML (SapientML) | LI/E | -1062.7 | -889.83 | 0.7484 | -137338 | -3.9674 | -19712 | -78814 | -2.6196 | -743.14 |
|  | PR | -1085.2 | -660.25 | 0.7228 | -0.5162 | -33.290 | -8036.7 | -90487 | -7.3229 | -688.38 |
|  | ES | -1061.9 | -392.37 | 0.7493 | -0.4039 | -31.634 | -9105.5 | -58587 | -5.1570 | -621.04 |
|  | KNN | -1463.1 | -527.27 | 0.7691 | -0.4854 | -33.212 | -4793.2 | -60269 | -0.5136 | -1056.8 |
|  | None | -23018.9 | -111678 | -10.389 | -496.44 | -30474.9 | -51271.6 | -1144894 | -4393.94 | -18723.7 |

Note: Humi = Humidity, Temp = Temperature, Pres = Pressure

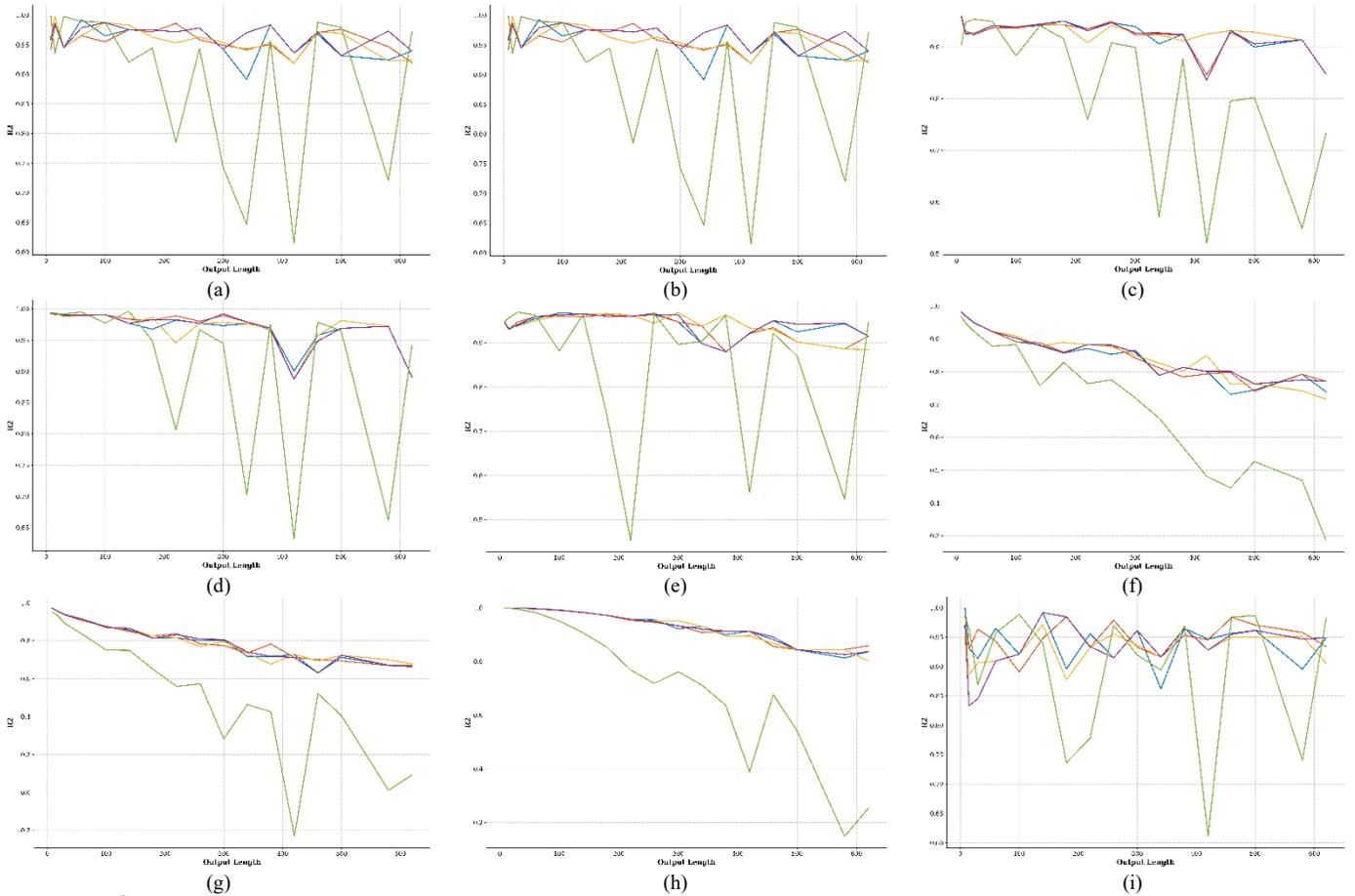

**Fig. 6.** The $R^2$ chart for output lengths with each estimator for data center dataset (Blue: LI/E, Red: PR, Yellow: ES, Purple: KNN, Green: None): (a) $PM_1$ – XGBoost; (b) $PM_{10}$ –XGBoost, (c) $P_{0.3}$ – XGBoost, (d) $P_{0.5}$ – XGBoost, (e) $P_{1.0}$ – XGBoost, (f) humidity – LightBGM, (g) temperature – LightBGM, (h) pressure – LightBGM, (i) AQI – XGBoost.

As shown in Fig. 6, this study presents an $R^2$ chart depicting the performance of various estimators across different output lengths on the dataset experiment. To maintain conciseness, the evaluation exclusively showcases the ATLO-ML models that demonstrate the highest $R^2$ value for each feature. Specifically, the $R^2$ values are displayed for LightGBM predicting humidity, temperature, and pressure; XGBoost predicting $PM_{10}$, $P_{0.3}$, $P_{0.5}$, $P_{1.0}$ and AQI. The visualization uses distinct colored lines to show each estimator's performance: blue for Linear Interpolation/Extrapolation, orange for PR, yellow for ES, purple for KNN Regression, and green for cases without an estimator.

As shown in Fig. 6, the same phenomenon is within Fig. 4. While extending the prediction time length, the $R^2$ value decreases. In this dataset, the analysis observed that the prediction accuracy was relatively lower compared to the GAM-dataset. Without implementing ATLO-ML, the $R^2$ values exhibited significant fluctuations across different time periods, resulting in non-linear relationships with the measured outcomes. However, when ATLO-ML was employed, the predictive performance demonstrated marked improvement over the non-ATLO-ML approach, with $R^2$ values showing a more gradual decline as the temporal window expanded. The findings indicate that optimal prediction accuracy for various time horizons necessitates the calibration of distinct sampling rates and input lengths. ATLO-ML serves as an automated algorithmic solution that assists users in identifying these optimal parameters.

Among these parameters, AQI exhibits particularly noteworthy characteristics. In AQI predictions, the results showed that the $R^2$ values did not demonstrate the typical declining trend associated with extended time horizons. Instead, they maintained relatively stable values within a consistent range. However, without ATLO-ML implementation, this range exhibited greater variability compared to scenarios where ATLO-ML was applied. This indicates that ATLO-ML achieves superior stability in predictive performance compared to conventional machine learning approaches.

The experimental results indicate that AutoML, despite incorporating the ATLO-ML structure, fails to demonstrate consistent and superior performance compared to conventional methodologies. Upon investigation, the analysis identified that this limitation stems from AutoML's primary design focus on tabular data rather than time series analysis. Consequently, when faced with time series datasets characterized by higher levels of uncertainty and complexity, AutoML exhibits limited capability in effectively analyzing column semantics and generating appropriate preprocessing pipelines and training models.

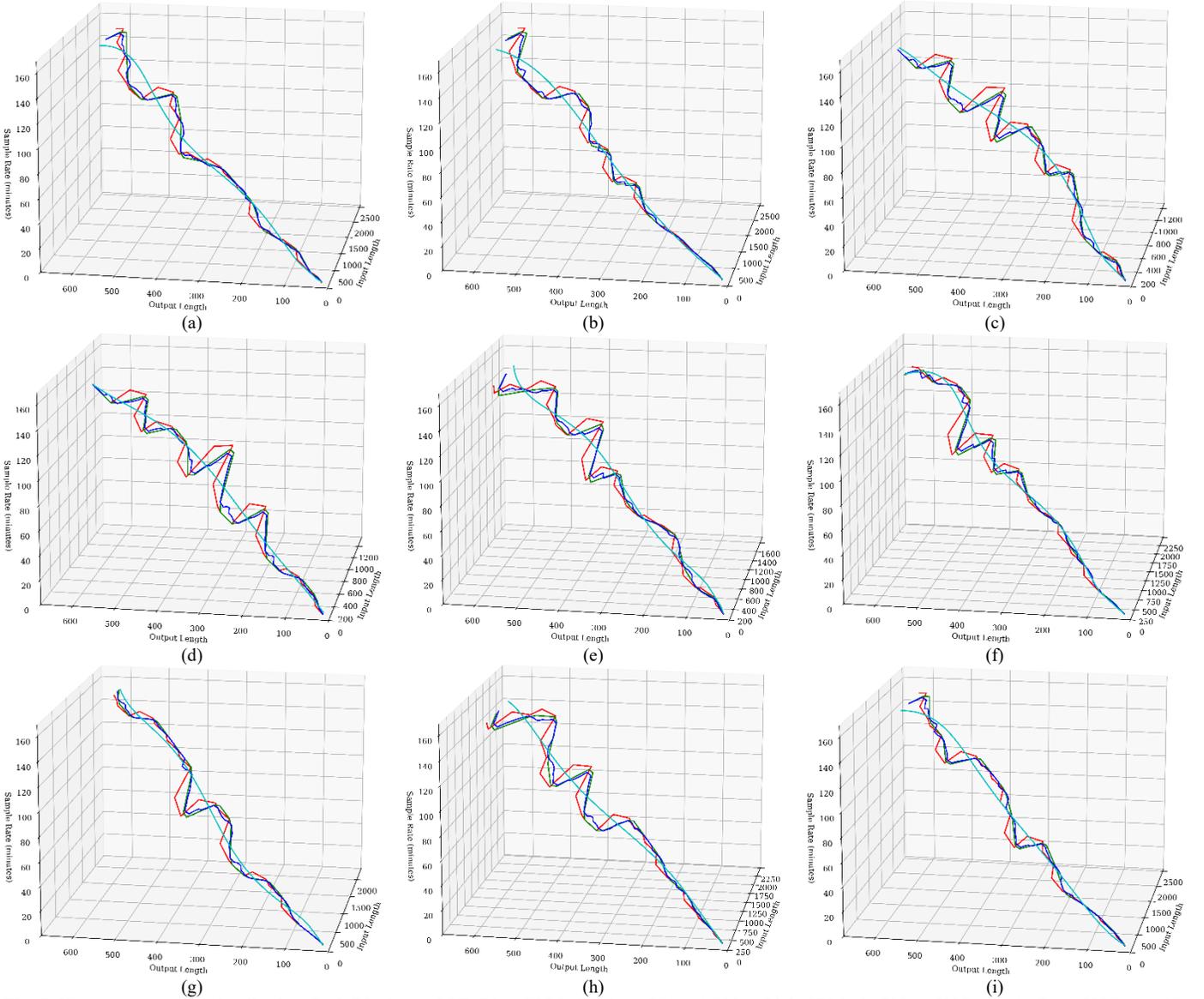

**Fig. 7.** 3D parameter space visualization (Red: ES, Green: LI/E, Blue: KNN, Cerulean: PR): (a) $PM_1$ – LightGBM; (b) $PM_{10}$– XGBoost; (c) $P_{0.3}$ – XGBoost; (d) $P_{0.5}$– XGBoost; (e) $P_{1.0}$– XGBoost; (f) Humidity – LightGBM, (g) Temperature – LightGBM, (h) Pressure – LightGBM, and (i) AQI – XGBoosts

Fig. 7 demonstrates the variations in parameter space. Due to space limitations, the parameter space presentation is limited to the best-performing machine learning model for each feature. As shown in Fig. 5 and 7, the parameter space curves trained from both databases exhibit striking similarities. Despite slight variations in proportions among input length, output length, and sample rate, the overall trajectory of the curves remains consistent. This observation indicates that when utilizing machine learning for time series prediction, the input length and sample rate must be proportionally fine-tuned according to the output length during optimization to achieve optimal results. However, this process typically requires extensive manual adjustment. This phenomenon reinforces the necessity of ATLO, reducing the demand for manual parameter space tuning.

*C. Experimental Components*

The experiments were conducted using a high-performance computing environment. The hardware configuration consisted of an Intel® Xeon® CPU E5-2697A v4 processor operating at 2.60GHz, complemented by 441GB of system memory. The software environment was based on Ubuntu 22.04.5 LTS operating system.

For the implementation of the proposed methods, the study utilized a suite of state-of-the-art machine learning libraries and frameworks. Specifically:

- Python version 3.10.12 served as the primary programming language.
- SapientML version 0.4.15 was employed for automated machine learning tasks.
- Scikit-Learn version 1.3.2 provided a comprehensive set of tools for data preprocessing and model evaluation.
- LightGBM version 4.5.0 and XGBoost version 2.1.1 were used for gradient boosting implementations.
- Matplotlib version 3.9.2 is used to ensure consistent and high-quality graphical representations of the findings.

This configuration enabled us to perform robust computational experiments and analyses, ensuring the reliability and reproducibility of the results.

## V. Conclusion

This paper introduces ATLO-ML, an adaptive time-length optimization system that automatically determines optimal input time length and sampling rate parameters for time series prediction in air quality monitoring. The experimental results demonstrate that ATLO-ML significantly improves prediction accuracy compared to fixed parameter approaches across multiple machine learning models and environmental parameters. Through comprehensive validation using both public GAMS-dataset and proprietary data center measurements, the system showed remarkable ability to mitigate accuracy degradation over extended prediction horizons. The research revealed that different air quality parameters require distinct input length and sampling rate ratios, challenging the conventional practice of using fixed ratios. ATLO-ML's estimators consistently outperformed baseline approaches, with particularly strong performance in LightGBM and XGBoost implementations, demonstrating the framework's effectiveness in optimizing temporal parameters for machine learning workflows. It is important to note that while these findings highlight the potential of ATLO-ML, the optimal choice of estimator may still vary depending on specific dataset characteristics and problem domains.

## VI. Future Work

For future research directions, several promising avenues warrant investigation. Further exploration could focus on understanding the underlying factors that contribute to estimator performance variations across different environmental parameters, potentially leading to more targeted optimization strategies. Integration possibilities between ATLO-ML and existing AutoML frameworks could yield enhanced capabilities for time series prediction. Additionally, extending the system's validation to diverse domains beyond air quality monitoring would help establish its broader applicability and robustness. Research efforts could also focus on optimizing the parameter space exploration process to reduce computational overhead while maintaining prediction accuracy and developing adaptive mechanisms for automatic estimator selection based on specific dataset characteristics and problem domains. These advancements would further strengthen ATLO-ML's position as a versatile solution for time series parameter optimization across various applications.

## Acknowledgment

The authors thank Wenli Yu, the chief executive officer of Archimedes Controls Corp., for providing the IoT system A150 and access to the data center as an experimental site.

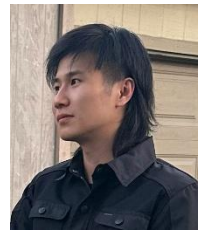

**I-Hsi Kao** was born in Taipei, Taiwan, in 1994. He earned his B.S. degree in Electrical Engineering from National United University, Miaoli, Taiwan, in 2016, followed by M.S. and Ph.D. degrees in Mechanical and Electromechanical Engineering from National Sun Yat-sen University, Kaohsiung, Taiwan, in 2018 and 2021, respectively.

He was a research assistant at National Sun Yat-sen University in Kaohsiung, Taiwan, from 2016 to 2020. He served as a Private in the Taiwan (ROC) Army in 2021. From 2020 to 2022, he was a Visiting Scholar at the University of California, Berkeley, CA, USA, followed by a Postdoctoral Researcher role from 2022 to 2023. In 2023, he joined Fujitsu Research of America Inc., Santa Clara, CA, USA, where he currently serves as a Tech Lead and Architecture. His research interests include Artificial Intelligence, Intelligent Sensing, Intelligent Transportation, and Systems Engineering.

Dr. Kao was a Guest Editor at *Sensors MDPI* leading special issue of *The Intelligent Sensing Technology of Transportation System* in 2022. He won the Best Paper Awards at *IEEE International Conference on Consumer Electronics-Taiwan* in both 2021 and 2023. He has served as Chair of *Electrical Engineering / Computer Sciences* at the *Chinese Institute of Engineers / USA San Francisco Bay Area Chapter* since 2024. He has also served as a member of the *Machine Learning, Deep Learning, and AI in Consumer Electronics Technical Committee* within the *IEEE Consumer Technology Society* since 2024.

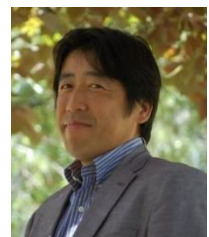

**Kanji Uchino** was born in Nagoya, Japan, in 1968. He earned his B.S. degree in information science from University of Tsukuba, Tsukuba, Japan, in 1991, followed by M.S. and Ph.D. degrees in science and engineering, information engineering from University of Tsukuba, in 1993 and 1996, respectively.

He joined Fujitsu Laboratories LTD., Kawasaki, Japan, in 1996. He is currently a Director of Research at Fujitsu Research of America Inc., Santa Clara, CA, USA. His research interests include artificial intelligence (Trusted AI, Graph AI, AutoML, etc.), open education, and open innovation.

Dr. Uchino has been a member of Information Processing Society of Japan (IPSJ) since 1992, The Japanese Society for Artificial Intelligence (JSAI) since 1996.